\documentclass[review]{elsarticle}

\usepackage{lineno,hyperref}
\usepackage{amsmath,amssymb,amsfonts}
\usepackage{enumitem}
\usepackage{multirow}
\usepackage{graphicx}
\usepackage{caption}
\usepackage{subcaption}
\modulolinenumbers[5]

\begin{document}

\begin{frontmatter}

\title{Towards in-store multi-person tracking using head detection and track heatmaps}
\tnotetext[mytitlenote]{This work was done and completed for Sunmi US Inc.}





\author[1]{Aibek Musaev\corref{cor1}}
\author[1]{Jiangping Wang}
\author[2]{Liang Zhu}
\author[2]{Cheng Li}
\author[1]{Yi Chen}
\author[1]{Jialin Liu}
\author[2]{Wanqi Zhang}
\author[2]{Juan Mei}
\author[1,2]{De Wang\fnref{fn2}}
\address[1]{Sunmi AI lab, Sunmi US Inc., Pleasanton, CA 94588}
\address[2]{Sunmi AI lab, Shanghai Sunmi Technology Co., Ltd, Shanghai, China, 200433}
\cortext[cor1]{Corresponding author1: Aibek Musaev (e-mail: aibek.musaev@sunmi.com)}
\cortext[cor2]{Corresponding author2: De Wang (e-mail: de.wang@sunmi.com)}

\begin{abstract}
Computer vision algorithms are being implemented across a breadth of industries to enable technological innovations. In this paper, we study the problem of computer vision based customer tracking in retail industry. To this end, we introduce a dataset collected from a camera in an office environment where participants mimic various behaviors of customers in a supermarket. In addition, we describe an illustrative example of the use of this dataset for tracking participants based on a head tracking model in an effort to minimize errors due to occlusion. Furthermore, we propose a model for recognizing customers and staff based on their movement patterns. The model is evaluated using a real-world dataset collected in a supermarket over a 24-hour period that achieves 98\% accuracy during training and 93\% accuracy during evaluation.
\end{abstract}


\end{frontmatter}


\section{Introduction}

Recent advancements in computer vision research on image classification, object detection, and object tracking enabled technological innovations in many industries including transportation~\cite{DBLP:conf/ijcnn/OnishiMSMO19}, healthcare~\cite{DBLP:conf/icdm/ZhangCW18}, and agriculture~\cite{DBLP:conf/cvpr/WuZLCY19}.

In this paper, we study the application of computer vision in retail industry. Specifically, our focus is on supporting in-store analytics based on customer tracking. Retailers are using such analytics to study buying patterns~\cite{DBLP:conf/ism/LiuGK15}, combat fraud~\cite{connell2013retail}, and estimate in-store traffic~\cite{8576169}.

The area of computer vision research that can be applied for the customer tracking problem is called multiple object tracking (MOT). MOT algorithms estimate the trajectory of different objects in a sequence which is usually a video. The standard approach in MOT algorithms is tracking-by-detection where a set of detections are extracted from the video frames and then used to guide the tracking process. A tracking algorithm associates such detections together by assigning the same ID to the detections that contain the same target.

There are multiple challenges affecting the accuracy of the MOT algorithms including occlusions and interactions between targets that can sometimes also have similar appearance. We propose to mount surveillance cameras on the ceiling and train MOT algorithms to use human heads for object detection as opposed to full bodies. This will help to reduce the effect of occlusions on the system performance.

Furthermore, the models built for customer tracking in retail industry should take into account additional behaviors that are specific to supermarkets, such as squatting when picking up products from lower shelves or standing in close proximity to one another when waiting in lines.

Note the importance of having robust ground truth datasets that help the researchers train, evaluate and compare their models. The presence of solid, publicly accessible datasets enable the progress in various research areas, including image classification via ImageNet~\cite{russakovsky2015imagenet}, object detection via Pascal VOC~\cite{Everingham15} and Microsoft COCO~\cite{DBLP:conf/eccv/LinMBHPRDZ14}, and object tracking via MOTChallenge~\cite{MOT16}.

Also note that in most retail stores staff members are also present in addition to customers. Hence, it becomes critical to be able to distinguish between customers and staff when analyzing video data from surveillance cameras. Furthermore, being able to identify staff members may enable a richer set of analytical reports for retailers. For example, the following automated reports may become possible:

\begin{itemize}
    \item number of staff members in a store at any given moment;
    \item for each staff member, how many hours are spent in a store;
    \item distribution of time spent at a cashier's desk versus elsewhere;
    \item which employee opened/closed the store, and so on.
\end{itemize}

After analyzing the actual videos from surveillance cameras, we observe that customers and staff have distinct movement patterns. For example, in a supermarket, staff members spend more time behind a cashier's desk whereas customers tend to browse the store aisles and stop by in front of a cashier's desk when they are ready to check out. This observation leads to the following hypothesis: it may be possible to classify people appearing in videos obtained from in-store cameras as either customers or staff based on their movement patterns.


\subsection{Research challenges}

There are several challenges in multiple object tracking (MOT). The main difficulties in tracking multiple targets simultaneously include various occlusions and interactions between objects that can sometimes also have similar appearance~\cite{DBLP:journals/ijon/CiaparroneSTTTH20}.

The standard approach employed in MOT algorithms is a two-step process: a set of detections are first extracted from video frames as bounding boxes using object detection; then track IDs are assigned to those bounding boxes using object tracking. Note that both object detection and tracking algorithms are imperfect and contribute errors to the system's performance.

Lastly, it is difficult to recognize a person in a video as either a customer or a staff member if the staff do not wear a uniform. In fact, some of the authors of this work either were at some point mistakenly perceived to be staff members in a supermarket or perceived other customers as staff members. The problem is amplified when such recognition is to be performed by a computer algorithm as opposed to a human.

\subsection{Our contributions}

In summary, our key contributions are:

\begin{itemize}
    \item \textbf{Fully annotated in-office dataset.} We release the annotated dataset of participants in an office environment exhibiting six different behaviors that are typical for supermarkets\footnotemark. For each video sequence, 3 types of files are provided: image files in .JPEG format, annotations in .XML format, and video files in .MP4 format.
    \item \textbf{Illustrative example of the use of the in-office dataset.} We evaluate the proposed head tracking model for multiple person tracking using a combination of object detection and object tracking algorithms. The effect of the errors contributed by these algorithms on the overall system performance is evaluated individually as well as collectively. Furthermore, the effect of skipping video frames on the system performance is also evaluated.
    \item \textbf{Customer and staff recognition based on track heatmaps.} We propose a model for customer and staff recognition in a supermarket environment based on their movement patterns. The model uses a two-step process as follows: 1) trajectories are used to generate heatmaps, 2) heatmaps are used as representation for learning a classification model. The model achieves a 93\% test accuracy based on a sample of real-world data collected in a store.
    \item \textbf{Sample annotated supermarket dataset.} We also release the real-world dataset collected in a supermarket store\footnotemark[\value{footnote}]\footnotetext{https://github.com/Sunmi-AI-Lab/head-detection-and-tracking}. This dataset contains the manually annotated tracks of customers and staff members based on a sample of a 24-hour video.
\end{itemize}

\section{Related work}

\subsection{Image classification}
\label{sec:image_classification}

In computer vision, image classification is a problem of producing a list of object categories present in the image such as humans or animals. Due to the importance of this problem, an annual competition called the ImageNet Large-Scale Visual Recognition Challenge (ILSVRC) has been held every year since 2010~\cite{russakovsky2015imagenet}. The challenge is based on a "trimmed" version of the ImageNet dataset containing 1.2 million images spread over 1,000 categories~\cite{deng2009imagenet}. In 2012, a breakthrough result has been achieved by Krizhevsky, et al.~\cite{krizhevsky2012imagenet}. A deep convolutional neural network (CNN) called AlexNet reduced classification error rate from 26\% to 15.3\%. This was one of the first deep convolutional neural networks to achieve considerable accuracy in this challenge. In 2014, the winner of this competition was GoogLeNet that reduced classification error rate to 6.67\%! The proposed network implemented a novel element called "inception" which drastically reduced the number of network parameters. The runner-up at the same competition was VGG-16 developed by Simonyan and Zisserman~\cite{DBLP:journals/corr/SimonyanZ14a}. This model is appealing due to its uniform architecture. However, it consists of 138 million parameters which can be challenging to compute.

In 2015, a new SOTA result was obtained based on the idea of deep residual networks or ResNet~\cite{he2016deep}. It reduced the classification error rate further down to 3.57\% which beats the human-level performance on this dataset. In this project, we use an updated version of ResNet that improves the learning of deep residual networks through the use of
identity-based skip connections~\cite{he2016identity}.

\subsection{Object detection}
\label{sec:object_detection}

Detecting persons in a video stream is an object detection problem. Fast, robust object detection algorithms are fundamental to the success of next-generation video processing systems. Such systems are capable of detecting various types of objects, such as airplanes, bicycles, birds, cars, persons, trains,  and other objects~\cite{everingham2010pascal}. 

Recently, deep convolutional neural networks have made significant advancements in image classification~\cite{DBLP:journals/cacm/KrizhevskySH17} and object detection~\cite{ren2015faster}. Object detection is a more challenging task as it combines image classification with object localization. Given an image, an object detection algorithm produces one or more bounding boxes with the class label attached to each bounding box. Such algorithms are capable enough to deal with multi-class classification and localization as well as to deal with objects with multiple occurrences.

Modern systems approach object localization in either a multi-stage pipeline (RCNN~\cite{DBLP:conf/cvpr/GirshickDDM14}, Fast RCNN~\cite{DBLP:conf/iccv/Girshick15}, Faster RCNN~\cite{DBLP:journals/pami/RenHG017}) or in a single-shot manner (YOLO~\cite{DBLP:conf/cvpr/RedmonDGF16}, SSD~\cite{DBLP:conf/eccv/LiuAESRFB16}). The multi-stage pipeline approaches all share one feature in common: one part of their pipeline is dedicated to generating region proposals followed by a high quality classifier to classify those proposals. These methods are very accurate, but come at a high computational cost (low frame-rate); in other words, they are not well suited for real-time object detection. Hence, in this project we focus our attention on single-shot approaches.

An alternative way of doing object detection is by combining these two tasks into one network also known as single-shot detection. You Only Look Once (YOLO) is a popular algorithm among single-shot detector approaches for object detection~\cite{DBLP:conf/cvpr/RedmonDGF16}. YOLO divides every image into a grid of SxS and every grid predicts N bounding boxes and confidence for each class. So, SxSxN boxes are forecasted in total. Given a threshold, e.g. 30\%, most predictions can be filtered out. The current iteration of this model is called YOLOv3 which is the version used in this project~\cite{DBLP:journals/corr/abs-1804-02767}.

SSD also belongs in the single-shot detection family but runs a convolutional neural network on the input image only once and computes a feature map~\cite{DBLP:conf/eccv/LiuAESRFB16}. Then a small 3x3 sized convolutional kernel is run on this feature map to foresee the bounding boxes and class probabilities. The original SSD paper used the VGG-16 model as its convolutional neural network~\cite{DBLP:journals/corr/SimonyanZ14a}. In this work, VGG-16 is replaced by the ResNet-50 model described in Section~\ref{sec:image_classification}.

\subsection{Multiple Object Tracking (MOT)}

Multiple object tracking (MOT) is a computer vision task whose goal is to estimate the trajectory of different objects in a sequence which is usually a video. Compared to object detection, whose output is a collection of rectangular bounding boxes identified by their coordinates, height, and width, MOT algorithms also associate a target ID to each box. Recently, the algorithms that provide a solution to this problem have benefited from the rise of deep learning models~\cite{DBLP:journals/ijon/CiaparroneSTTTH20}. This led to the creation of the MOTChallenge in 2015 which is a benchmark whose goal is to collect
existing and new data and create a framework for the standardized evaluation of multiple object tracking methods~\cite{MOTChallenge2015}. Since then, new versions of this benchmarks have been released with more challenging and diverse video sequences~\cite{MOT16, MOT19_CVPR, MOTChallenge20}.

The standard approach employed in MOT algorithms is tracking-by-detection: a set of detections are extracted from the video frames and are used to guide the tracking process~\cite{DBLP:journals/ijon/CiaparroneSTTTH20}. A tracking algorithm associates such detections together by assigning the same ID to bounding boxes that contain the same target. Modern detection frameworks described in Section~\ref{sec:object_detection} ensure a reasonable detection quality. In this project, both YOLOv3 and SSD ResNet-50 detection models are used to compute the bounding boxes.

In 2015, a simple online and real-time tracking algorithm called SORT was ranked the best open source multiple object tracker in the MOTChallenge~\cite{DBLP:conf/icip/BewleyGORU16}. Despite only using a combination of familiar techniques such as the Kalman Filter and Hungarian algorithm for the tracking components, this approach achieved an accuracy comparable to SOTA trackers while maintaining high processing speed. In this project, we use an improved version of this algorithm called Deep SORT~\cite{DBLP:conf/icip/WojkeBP17, DBLP:conf/wacv/WojkeB18}. It extends the original SORT algorithm by integrating appearance information based on a deep appearance descriptor.

The CLEAR MOT metrics were developed for the Classification of Events, Activities and Relationships (CLEAR) workshops held in 2006~\cite{DBLP:conf/clear/2006} and 2007~\cite{DBLP:conf/clear/2007}. Those metrics include MOTA (Multiple Object Tracking Accuracy) that serves as a summary of other simpler metrics which compose it. The simpler metrics are the following:

\begin{itemize}
    \item FP: the number of false positives in the whole video;
    \item FN: the number of false negatives in the whole video;
    \item IDSW: the total number of ID switches;
    \item GT: the number of ground truth boxes.
\end{itemize}

The MOTA score is then defined as follows:

\begin{equation*}
    MOTA=1-\frac{FN+FP+IDSW}{GT} \quad \in (-\infty,1].
\end{equation*}

In this project, we use MOTA to evaluate the performance of various MOT algorithms.

\section{In-office dataset description}

This dataset is generated using a video that was shot from a single camera in an office environment. The camera was mounted to the ceiling. Multiple participants were involved in the video where they mimicked six different behaviors that are typical for customers in a supermarket as follows:

\begin{itemize}
    \item \textit{wearing hats}: participants are wearing hats;
    \item \textit{covering heads}: participants are using hands to cover their heads;
    \item \textit{squatting}: participants are squatting as if picking up products from lower shelves;
    \item \textit{putting jackets on/off}: participants are putting their jackets on and off;
    \item \textit{lights off}: lights are turned off while participants continue moving;
    \item \textit{close proximity}: participants are maintaining a close proximity to one another with possibly overlapping heads.
\end{itemize}

See Table~\ref{table:dataset} for an overview of the in-office dataset. It shows the total number of frames for each video sequence as well as the number of annotated frames.

For each video sequence, 3 types of files are provided: image files in .JPEG format, annotations in .XML format, and the original videos in .MP4 format.

\begin{figure*}[t]

\begin{subfigure}{0.32\textwidth}
\includegraphics[width=0.9\linewidth]{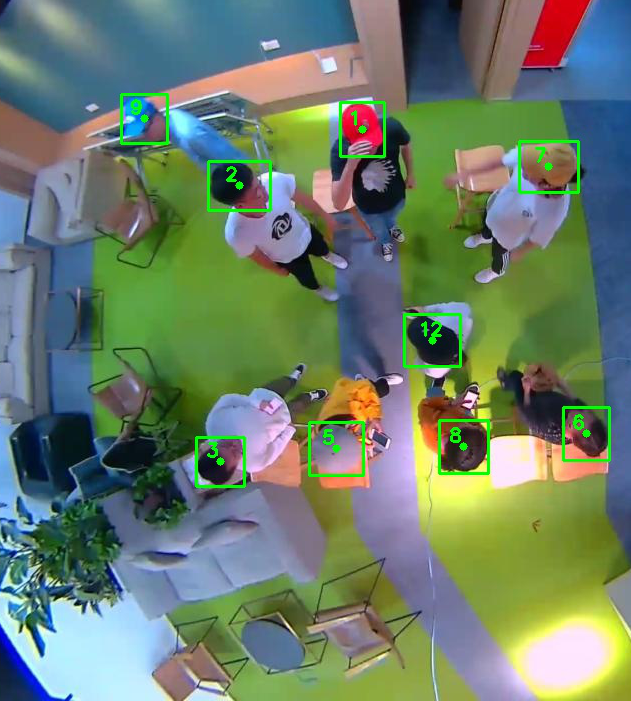}
\caption{Wearing hats}
\end{subfigure}
\begin{subfigure}{0.32\textwidth}
\includegraphics[width=0.9\linewidth]{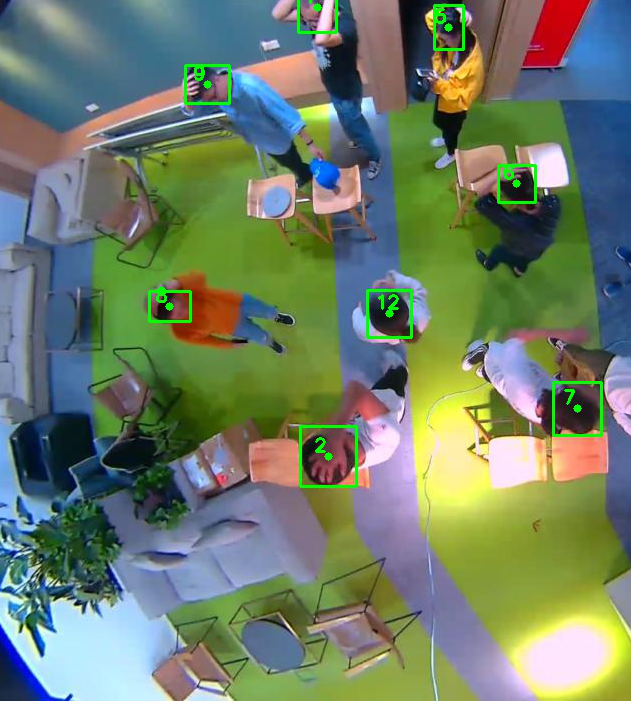}
\caption{Covering heads}
\end{subfigure}
\begin{subfigure}{0.32\textwidth}
\includegraphics[width=0.9\linewidth]{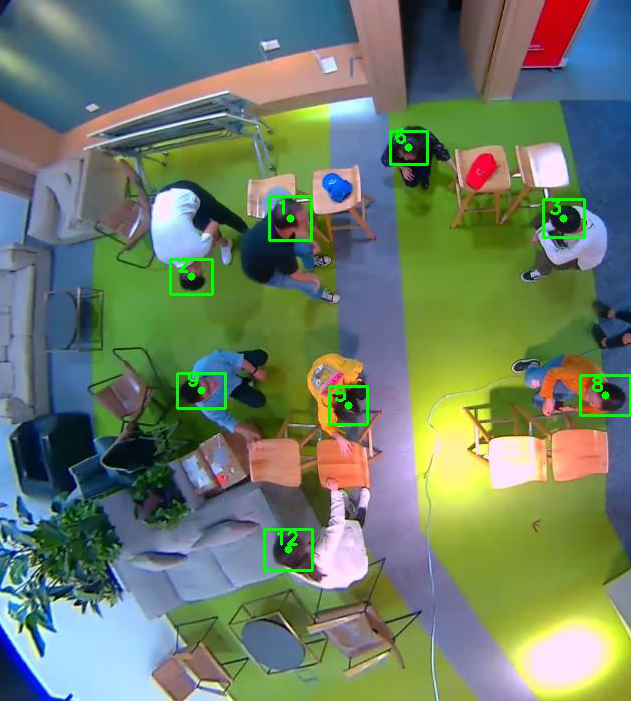}
\caption{Squatting}
\end{subfigure}
\begin{subfigure}{0.32\textwidth}
\includegraphics[width=0.9\linewidth]{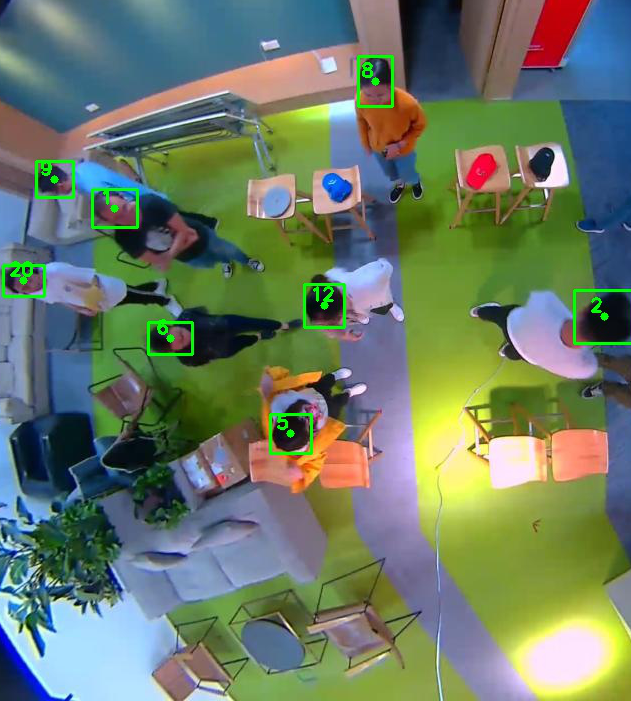}
\caption{Putting jackets on/off}
\end{subfigure}
\begin{subfigure}{0.32\textwidth}
\includegraphics[width=0.9\linewidth]{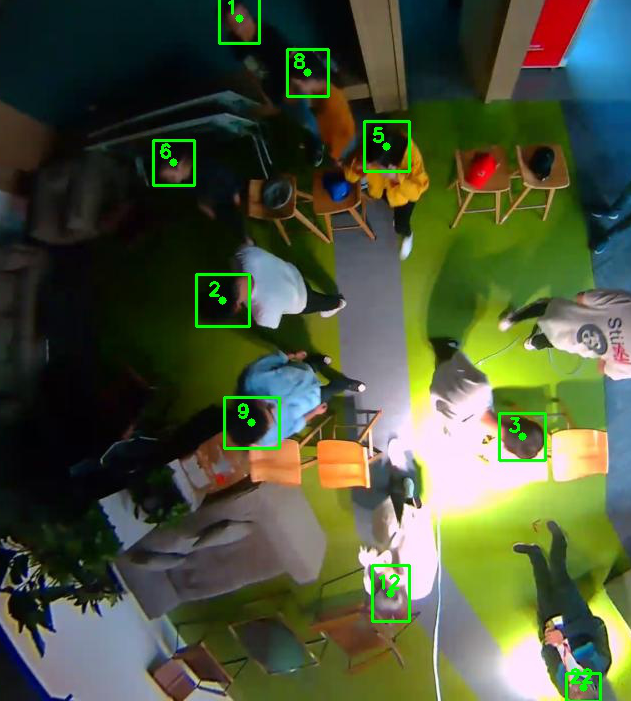}
\caption{Lights off}
\end{subfigure}
\begin{subfigure}{0.32\textwidth}
\includegraphics[width=0.9\linewidth]{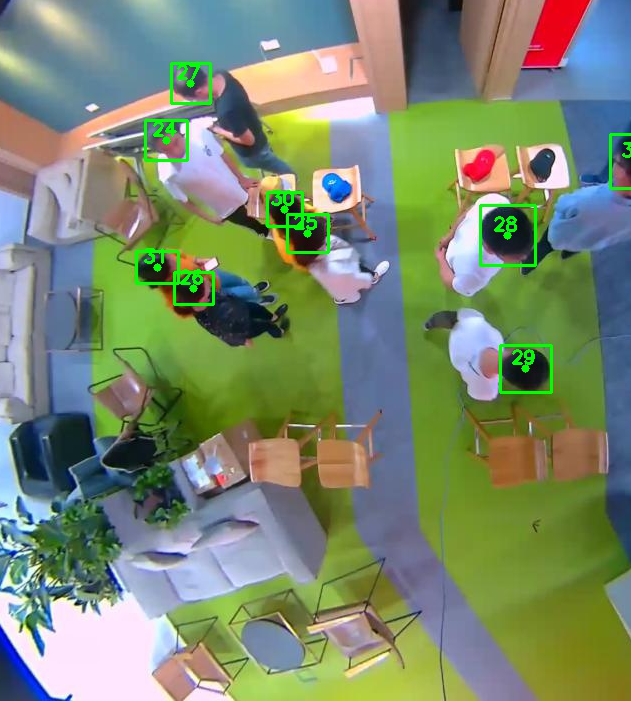}
\caption{Close proximity}
\end{subfigure}

\caption{Sample images from the in-office dataset for each type of behavior}
\label{fig:sample_images}
\end{figure*}

We show a sample image for each type of behavior in Figure~\ref{fig:sample_images}. Each image shows a bounding box around each head in the image and its track ID.

Note that a given person may have multiple track IDs assigned to her in this dataset. This is due to the following annotation strategy that was applied in this project. Anytime a person leaves the field-of-view (FOV), we consider it as the end of her current track. Similarly, anytime a person enters or re-enters the FOV, a new track ID is assigned to her. An alternative annotation strategy would be to assign the same track ID to the person re-entering the FOV in a given video. However, we did not explore this strategy in this project.

\begin{table}
\begin{tabular}{ |l|c|c|c| }
 \hline
 \textbf{Behavior} & \textbf{Video size} & \textbf{\# of frames} & \textbf{\# of annotations} \\
 \hline
 Wearing hats & 31.3 MB & 2,739 & 165 (6\%)\\
 Covering heads & 9.7 MB & 817 & 817 (100\%) \\
 Squatting & 11.1 MB & 900 & 167 (18\%)\\
 Putting jackets on/off & 17.1 MB & 1,447 & 166 (11\%)\\
 Lights off & 15.5 MB & 1,342 & 145 (10\%)\\
 Close proximity & 12.2 MB & 1,052 & 167 (15\%)\\
 \hline
\end{tabular}
\caption{Overview of the in-office dataset of participants in an office environment with six different behaviors}
\label{table:dataset}
\end{table}




\section{Recognizing customers and staff based on trajectory heatmaps}

Recognizing customers and staff based on their trajectories is not trivial. Recall that trajectories are represented by track IDs that are associated with a series of bounding boxes across a sequence of video frames. Note that bounding boxes are generated by an object detection algorithm while track IDs are generated as a result of an object tracking algorithm. Each of these algorithms are imperfect and have a certain error rate. This means that the output tracks may be erroneous as well.

Note that classifier algorithms expect numerical data which is typically provided in a vector or matrix form of fixed size. However, each track consists of a varying number of video frames that it appears in. So, the challenge is to come up with a meaningful representation of tracks of varying length as vectors or matrices of fixed length.

In this project, we propose a two-step process for converting trajectories of varying length to a fixed size representation as follows: 1) generate heatmaps  based on trajectories, 2) use the heatmap images as representation for learning a classification model.

\textbf{Step 1: Heatmap generation.} A heatmap image's size is set to the video frame size. Next, we accumulate exposure time per pixel for each object of interest, i.e. the participants' heads as we use a head tracking model. Each head is represented as a circle whose center coincides with the center of the corresponding bounding box.

\textbf{Step 2: Image classification.} We use an existing image classification approach, namely a popular ResNet-50 model, which is described in Section~\ref{sec:image_classification}. The classification model is built using the heatmap images.

\begin{figure*}[t]
\centering
\begin{subfigure}{0.32\textwidth}
\includegraphics[width=0.9\linewidth]{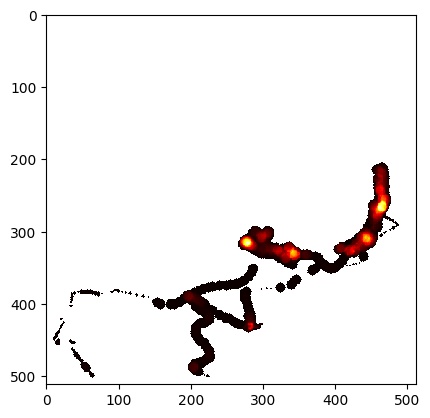}
\caption{Trajectory of a customer}
\end{subfigure}
\begin{subfigure}{0.32\textwidth}
\includegraphics[width=0.9\linewidth]{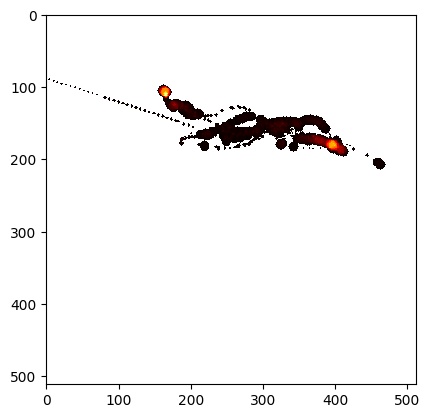}
\caption{Trajectory of a staff member}
\end{subfigure}
\begin{subfigure}{0.32\textwidth}
\includegraphics[width=0.9\linewidth]{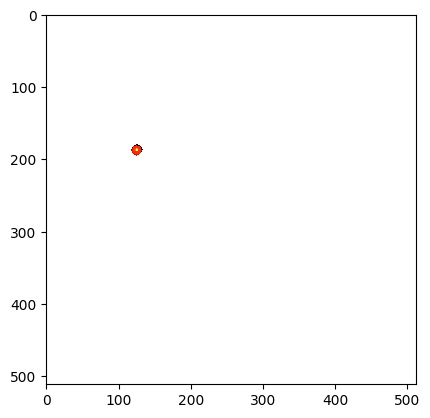}
\caption{Erroneous trajectory}
\end{subfigure}

\caption{Examples of heatmaps generated using trajectories of people in a supermarket}
\label{fig:sample_heatmaps}
\end{figure*}

See the examples of trajectory heatmaps in Figure~\ref{fig:sample_heatmaps}. It shows several examples of heatmaps generated using trajectories of people in a supermarket as well as an example of an erroneous trajectory heatmap. The staff member's trajectory heatmap shows that she spent most of her time at a cashier's desk located in the upper middle part of the image while the customer's trajectory is concentrated around the specific aisles in the store.

The erroneous trajectories are frequently a result of an inanimate object whose appearances are similar to a human's head. So, their trajectory's heatmap looks like a bright circle as shown in Figure~\ref{fig:sample_heatmaps}.

\section{Evaluation using real data}

\subsection{Experiment setup}

We conduct two sets of experiments using separate datasets. The first set of experiments is based on the in-office dataset. Specifically, we evaluate the effect of the errors of the individual components, namely object classification and object tracking as well as skipped frames, on the overall accuracy of the head tracking algorithm.

The second set of experiments is based on the real-world dataset collected in a supermarket store located in China. We apply the head tracking model to compute all tracks in this dataset. Then we generate a statistical sample of all the tracks, annotate it and evaluate the accuracy of the proposed customer versus staff classification.

\subsection{Evaluation of head tracking model using in-office dataset}
In this experiment, we evaluate the effect of the errors contributed by the individual components on the overall accuracy of the head tracking algorithm. The individual components are object detection and object tracking. Both of these components are not perfect and add errors to the overall performance. Our goal is to estimate the errors contributed by each of these components individually, as well as the compound error by both components.

In addition, we also evaluate the effect of skipping video frames on the overall performance. We start by skipping 10\% of the frames which means that 90\% of frames are remaining in the dataset. The frames to skip are computed based on a given probability such as 10\%. This is followed by skipping 20\% of the frames meaning that 80\% of the frames are remaining in the dataset. And so on, until 90\% of the frames are skipped resulting in only 10\% of the remaining frames.

Here is a summary of the evaluation combinations used in this set of experiments:

\begin{enumerate}
\item Effect of detection errors
    \begin{itemize}
        \item Test set. $bboxes_{system}$ : $tracks_{gt}$
        \item Ground truth set. $bboxes_{gt}$ : $tracks_{gt}$
    \end{itemize}
\item Effect of tracking errors
    \begin{itemize}
        \item Test set. $bboxes_{gt}$ : $tracks_{system}$
        \item Ground truth set. $bboxes_{gt}$ : $tracks_{gt}$
    \end{itemize}
\item Compound effect of tracking and detection errors
    \begin{itemize}
        \item Effect of detection errors. $bboxes_{system}$ : $tracks_{true}$
        \item Effect of tracking errors. $bboxes_{gt}$ : $tracks_{system}$
        \item Compound effect. $bboxes_{system}$ : $tracks_{system}$
    \end{itemize}
\end{enumerate}

These combinations will be explained in detail next.


\subsubsection{Effect of detection errors}
\label{sec:effect_of_detection_errors}

In the first experiment, we used two commonly used object detectors, namely YOLOv3 and SSD ResNet-50, for generating two sets of bounding boxes around heads in the video sequence. The bounding boxes are generated by the system, so we refer to them as ${bboxes_{system}}$. Then we manually annotated the tracks associated with those bounding boxes for each of these algorithms. The tracks are manually annotated and considered true, so we refer to them as ${tracks_{gt}}$. This means that the overall performance of the head tracking algorithm should only depend on the accuracy of the object detection component since the track IDs are true, i.e. correct.

For the ground truth set, we first manually draw the bounding boxes around each head (${bboxes_{gt}}$) and then manually annotate each head's track associated with the bounding boxes (${tracks_{gt}}$). This allows us to evaluate the effect of the object detection algorithm on the overall accuracy of the head tracking model.

\begin{figure}[t!]
    \centering
    \includegraphics[width=\linewidth]{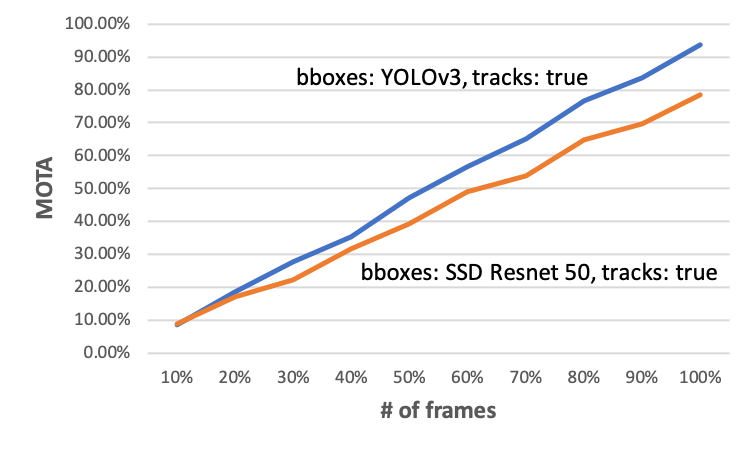}
    \caption{Effect of different detection algorithms on the overall accuracy of the head tracking algorithm.}
    \label{fig:exp_yolov3_vs_ssd}
\end{figure}

See the result of the experiment in Figure~\ref{fig:exp_yolov3_vs_ssd}. Both object detection algorithms demonstrate a linear correlation with respect to the number of frames in the dataset and the MOTA performance metric. However, YOLOv3 appears to outperform SSD ResNet-50 for the head tracking task on this dataset.

\subsubsection{Effect of tracking errors}
\label{sec:effect_of_tracking_errors}

In the second experiment, we use the manually drawn bounding boxes from the ground truth set, i.e. they are assumed to be true. Then we apply a DeepSORT tracking algorithm based on IOU matching strategy, i.e. without feature extraction, to generate tracks and compare it with a DeepSORT tracking algorithm with feature extraction.

\begin{figure}[t!]
    \centering
    \includegraphics[width=\linewidth]{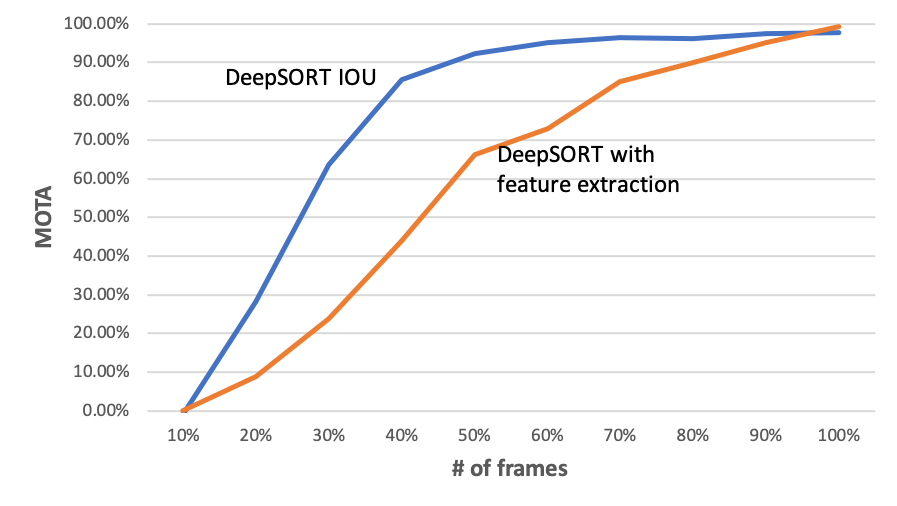}
    \caption{Effect of different tracking approaches on the overall accuracy of the head tracking algorithm.}
    \label{fig:exp_deepSORT_iou_vs_fe}
\end{figure}

See the result of the experiment in Figure~\ref{fig:exp_deepSORT_iou_vs_fe}. Although eventually DeepSORT with feature extraction outperforms a simpler model based on IOU matching strategy, it happens only when almost all video frames are preserved in the dataset. So, when skipping frames to speed up the processing pipeline, a simpler DeepSORT model may be sufficient for tracking purposes.

\subsubsection{Compound effect of detection and tracking errors}

The last experiment based on the in-office dataset evaluates the compound effect of the detection and tracking errors. Here, both head detection and head tracking are computed by the system. We compare the compound effect of these errors with the effect of errors in detection and the effect of errors in tracking components.

To compute the compound effect of detection and tracking errors, we use the bounding boxes generated by the head detection model based on YOLOv3 as it was shown to have a good performance compared to SSD ResNet-50. And then we use those detections as input to head tracking using DeepSORT IOU as it was shown to have a good performance compared to DeepSORT with feature extraction.

\begin{figure}[t!]
    \centering
    \includegraphics[width=\linewidth]{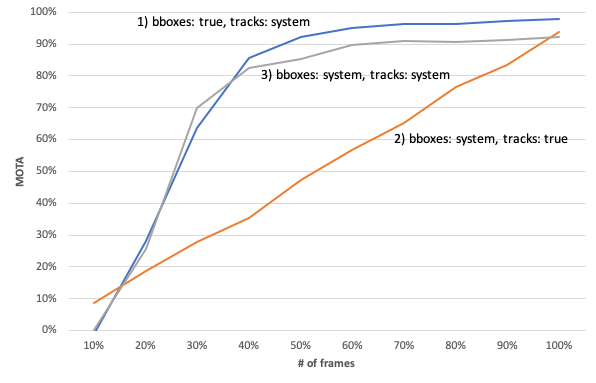}
    \caption{Compound effect of errors in detection and tracking on the overall accuracy of the head tracking algorithm.}
    \label{fig:exp}
\end{figure}

The system's performance with the errors in detection and the errors in tracking is described in Sections~\ref{sec:effect_of_detection_errors} and \ref{sec:effect_of_tracking_errors}. In this experiment, we plot the compound effect of errors in detection and tracking together with the effect of errors in detection and the errors in tracking in the same graph. See the result of the experiment in Figure~\ref{fig:exp}.

Note that the system's performance with errors in tracking outperforms the system's performance with the compound effect of errors in detection and tracking which is the expected behavior. However, the system's performance with the compound effect of errors unexpectedly outperforms the system's performance with errors in detection only. This is due to the behavior of the tracking algorithm which attempts to predict the missing frames in case of skipped frames. Thus, the resulting trajectories are closer to the ground truth with the true tracks that are missing frames.

\subsection{Customers and staff recognition based on trajectories}

In this experiment, we evaluate the accuracy of the algorithm that recognizes the customers and staff based on their trajectories. We use a 24-hour video shot from a camera located in a supermarket in China. The camera is mounted on the ceiling so we are able to apply our head tracking model for the generation of tracks.

See the overview of the collected dataset in Table~\ref{table:mozi}. We use YOLOv3 to generate bounding boxes~\cite{DBLP:journals/corr/abs-1804-02767} and Deep SORT to generate tracks using the collected bounding boxes~\cite{DBLP:conf/icip/WojkeBP17, DBLP:conf/wacv/WojkeB18}.

\begin{table}
\begin{tabular}{ |l|c|c|c|c| }
 \hline
 \textbf{Duration} & \textbf{Video size} & \textbf{\# of frames} & \textbf{\# of bboxes} & \textbf{\# of tracks} \\
 \hline
 24 hrs & 16.1 GB & 1,192,920 & 2,160,089 & 11,005\\
 \hline
\end{tabular}
\caption{Overview of the real-world dataset of participants in a supermarket environment collected over a 24-hour period}
\label{table:mozi}
\end{table}

\begin{figure}[t!]
    \centering
    \includegraphics[width=\linewidth]{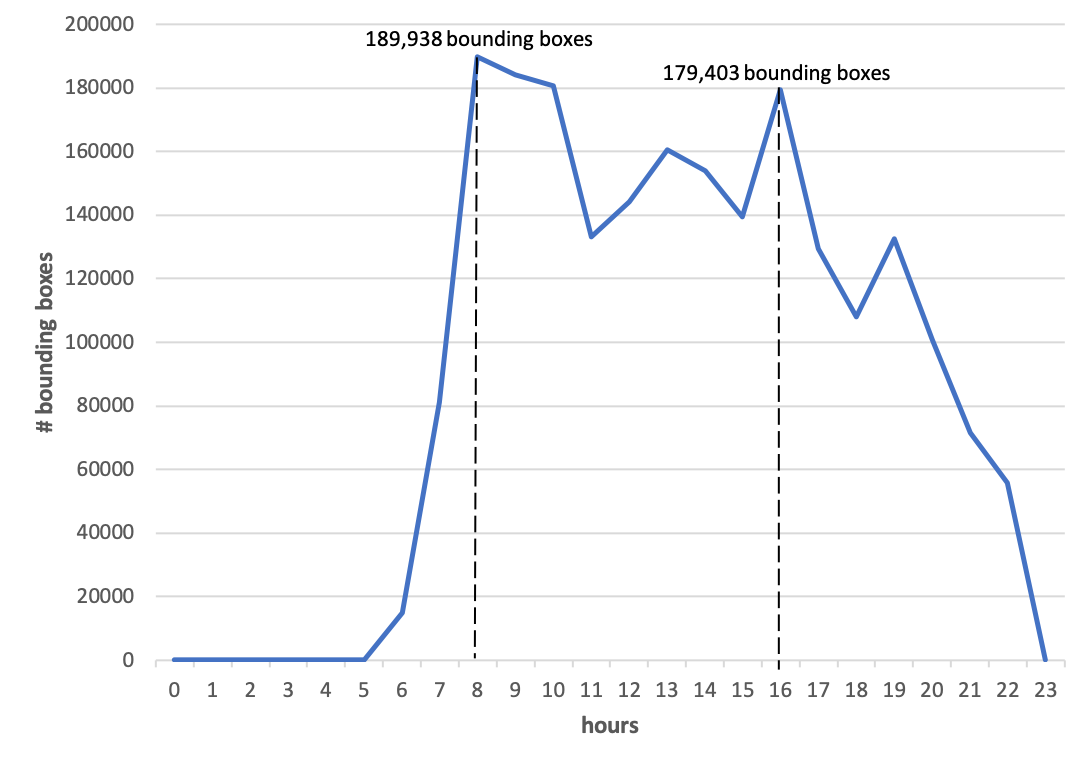}
    \caption{Overview of the unlabeled dataset by the number of bounding boxes detected per hour}
    \label{fig:dataset_unsupervised_bboxes}
\end{figure}

Figure~\ref{fig:dataset_unsupervised_bboxes} shows the store activity based on the number of bounding boxes detected per each hour in the video sequence. The peaks of those activities occur in the 8\textsuperscript{th} and 16\textsuperscript{th} hours.

\begin{figure}[t!]
    \centering
    \includegraphics[width=\linewidth]{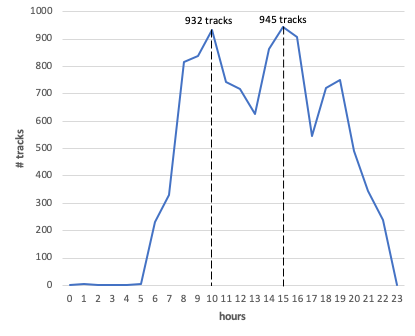}
    \caption{Overview of the real-world dataset by the number of tracks detected per hour}
    \label{fig:dataset_unsupervised_tracks}
\end{figure}

Figure~\ref{fig:dataset_unsupervised_tracks} shows the same store's activity using the number of tracks detected in each hour. Here, the peaks are slightly different, namely during the 10\textsuperscript{th} and 15\textsuperscript{th} hours.

As shown in Table~\ref{table:mozi}, there are 11,005 tracks in total. Given the population size of 11,005 tracks, 95\% confidence level, and 5\% margin of error, the sample size is calculated to be 313. Hence, 313 tracks were randomly sampled from the full set of tracks and then manually annotated based on the following classes:

\begin{itemize}
    \item \emph{0}: customer (36\%)
    \item \emph{1}: staff (33\%)
    \item \emph{2}: error (31\%)
\end{itemize}

Note that the percentage of errors in sampled tracks is roughly 1/3, which is extremely high.

Most of the tracks in the sample are short which may explain the high error rate in those tracks. Thus, the decision was made to preprocess tracks by only preserving sufficiently long tracks for analysis. But how do we define ``sufficiently long''? Let's define ``long'' first. There are at least two definitions of ``long'' with respect to tracks: either long in terms of the number of frames in a given track or long in terms of the distance covered in that track. We support both definitions as follows. We introduce the threshold for the acceptable number of frames per track which was empirically chosen to be $2,000$ frames for this video sequence. Whereas the threshold for the distance covered in a track is set to be at least twice the width of the video frame, i.e. $distance = image\_width * 2$.

The remaining number of tracks based on this filtering strategy is 920. Given the population size of 920, 95\% confidence level and 5\% margin of error, the sample size is calculated to be 272. Hence, 272 tracks were randomly sampled from the filtered set of tracks and then manually annotated based on the following classes:

\begin{itemize}
    \item 0: customer (49\%)
    \item 1: staff (49\%)
    \item 2: error (2\%)
\end{itemize}

So, there are only 2\% of errorneous tracks in the sample dataset compared to 31\% in the original dataset.

Next, we generate heatmaps for all tracks in the sample dataset, split them into train/validation/test datasets, build a classification model using ResNet-50, and evaluate its performance. The resulting \emph{\textbf{train accuracy is 98\%}} while the \emph{\textbf{test accuracy}} is \emph{\textbf{93\%}}. Hence, based on this real-world dataset the proposed model for customer and staff recognition shows a promising result that warrants further analysis.

\section{Conclusion}

In this paper, we address the problem of customer and staff member tracking using in-store ceiling cameras. The proposed model uses head tracking as opposed to full body recognition to reduce the effect of occlusions. The model is illustrated using the in-office dataset where participants exhibit various behaviors that are typical for supermarkets. Multiple object detection and tracking algorithms are compared and their performance is evaluated with respect to the skip-frame ratio, which is a common technique for maintaining a reasonable speed when processing videos. Furthermore, a model for customer and staff recognition in a supermarket environment based on their movement patterns is proposed. The model shows solid results using a sample of the real-world dataset collected in a supermarket over a 24-hour period. Lastly, all of the annotated datasets are released as a contribution to the research community.


\bibliography{mybibfile}

\begin{thebibliography}{10}
\expandafter\ifx\csname url\endcsname\relax
  \def\url#1{\texttt{#1}}\fi
\expandafter\ifx\csname urlprefix\endcsname\relax\def\urlprefix{URL }\fi
\expandafter\ifx\csname href\endcsname\relax
  \def\href#1#2{#2} \def\path#1{#1}\fi

\bibitem{DBLP:conf/ijcnn/OnishiMSMO19}
T.~Onishi, T.~Motoyoshi, Y.~Suga, H.~Mori, T.~Ogata,
  \href{https://doi.org/10.1109/IJCNN.2019.8852322}{End-to-end learning method
  for self-driving cars with trajectory recovery using a path-following
  function}, in: International Joint Conference on Neural Networks, {IJCNN}
  2019 Budapest, Hungary, July 14-19, 2019, {IEEE}, 2019, pp. 1--8.
\newblock \href {http://dx.doi.org/10.1109/IJCNN.2019.8852322}
  {\path{doi:10.1109/IJCNN.2019.8852322}}.
\newline\urlprefix\url{https://doi.org/10.1109/IJCNN.2019.8852322}

\bibitem{DBLP:conf/icdm/ZhangCW18}
X.~Zhang, J.~Chou, F.~Wang,
  \href{https://doi.org/10.1109/ICDM.2018.00092}{Integrative analysis of
  patient health records and neuroimages via memory-based graph convolutional
  network}, in: {IEEE} International Conference on Data Mining, {ICDM} 2018,
  Singapore, November 17-20, 2018, {IEEE} Computer Society, 2018, pp. 767--776.
\newblock \href {http://dx.doi.org/10.1109/ICDM.2018.00092}
  {\path{doi:10.1109/ICDM.2018.00092}}.
\newline\urlprefix\url{https://doi.org/10.1109/ICDM.2018.00092}

\bibitem{DBLP:conf/cvpr/WuZLCY19}
X.~Wu, C.~Zhan, Y.~Lai, M.~Cheng, J.~Yang,
  \href{http://openaccess.thecvf.com/content\_CVPR\_2019/html/Wu\_IP102\_A\_Large-Scale\_Benchmark\_Dataset\_for\_Insect\_Pest\_Recognition\_CVPR\_2019\_paper.html}{{IP102:}
  {A} large-scale benchmark dataset for insect pest recognition}, in: {IEEE}
  Conference on Computer Vision and Pattern Recognition, {CVPR} 2019, Long
  Beach, CA, USA, June 16-20, 2019, Computer Vision Foundation / {IEEE}, 2019,
  pp. 8787--8796.
\newblock \href {http://dx.doi.org/10.1109/CVPR.2019.00899}
  {\path{doi:10.1109/CVPR.2019.00899}}.
\newline\urlprefix\url{http://openaccess.thecvf.com/content\_CVPR\_2019/html/Wu\_IP102\_A\_Large-Scale\_Benchmark\_Dataset\_for\_Insect\_Pest\_Recognition\_CVPR\_2019\_paper.html}

\bibitem{DBLP:conf/ism/LiuGK15}
J.~Liu, Y.~Gu, S.~Kamijo, \href{https://doi.org/10.1109/ISM.2015.52}{Customer
  behavior recognition in retail store from surveillance camera}, in: 2015
  {IEEE} International Symposium on Multimedia, {ISM} 2015, Miami, FL, USA,
  December 14-16, 2015, {IEEE} Computer Society, 2015, pp. 154--159.
\newblock \href {http://dx.doi.org/10.1109/ISM.2015.52}
  {\path{doi:10.1109/ISM.2015.52}}.
\newline\urlprefix\url{https://doi.org/10.1109/ISM.2015.52}

\bibitem{connell2013retail}
J.~Connell, Q.~Fan, P.~Gabbur, N.~Haas, S.~Pankanti, H.~Trinh, Retail video
  analytics: an overview and survey, in: Video Surveillance and Transportation
  Imaging Applications, Vol. 8663, International Society for Optics and
  Photonics, 2013, p. 86630X.

\bibitem{8576169}
A.~{Generosi}, S.~{Ceccacci}, M.~{Mengoni}, A deep learning-based system to
  track and analyze customer behavior in retail store, in: 2018 IEEE 8th
  International Conference on Consumer Electronics - Berlin (ICCE-Berlin),
  2018, pp. 1--6.

\bibitem{russakovsky2015imagenet}
O.~Russakovsky, J.~Deng, H.~Su, J.~Krause, S.~Satheesh, S.~Ma, Z.~Huang,
  A.~Karpathy, A.~Khosla, M.~Bernstein, et~al., Imagenet large scale visual
  recognition challenge, International journal of computer vision 115~(3)
  (2015) 211--252.

\bibitem{Everingham15}
M.~Everingham, S.~M.~A. Eslami, L.~Van~Gool, C.~K.~I. Williams, J.~Winn,
  A.~Zisserman, The pascal visual object classes challenge: A retrospective,
  International Journal of Computer Vision 111~(1) (2015) 98--136.

\bibitem{DBLP:conf/eccv/LinMBHPRDZ14}
T.~Lin, M.~Maire, S.~J. Belongie, J.~Hays, P.~Perona, D.~Ramanan,
  P.~Doll{\'{a}}r, C.~L. Zitnick,
  \href{https://doi.org/10.1007/978-3-319-10602-1\_48}{Microsoft {COCO:} common
  objects in context}, in: D.~J. Fleet, T.~Pajdla, B.~Schiele, T.~Tuytelaars
  (Eds.), Computer Vision - {ECCV} 2014 - 13th European Conference, Zurich,
  Switzerland, September 6-12, 2014, Proceedings, Part {V}, Vol. 8693 of
  Lecture Notes in Computer Science, Springer, 2014, pp. 740--755.
\newblock \href {http://dx.doi.org/10.1007/978-3-319-10602-1\_48}
  {\path{doi:10.1007/978-3-319-10602-1\_48}}.
\newline\urlprefix\url{https://doi.org/10.1007/978-3-319-10602-1\_48}

\bibitem{MOT16}
A.~Milan, L.~Leal-Taix\'{e}, I.~Reid, S.~Roth, K.~Schindler,
  \href{http://arxiv.org/abs/1603.00831}{{MOT}16: {A} benchmark for
  multi-object tracking}, arXiv:1603.00831 [cs]ArXiv: 1603.00831.
\newline\urlprefix\url{http://arxiv.org/abs/1603.00831}

\bibitem{DBLP:journals/ijon/CiaparroneSTTTH20}
G.~Ciaparrone, F.~L. S{\'{a}}nchez, S.~Tabik, L.~Troiano, R.~Tagliaferri,
  F.~Herrera, \href{https://doi.org/10.1016/j.neucom.2019.11.023}{Deep learning
  in video multi-object tracking: {A} survey}, Neurocomputing 381 (2020)
  61--88.
\newblock \href {http://dx.doi.org/10.1016/j.neucom.2019.11.023}
  {\path{doi:10.1016/j.neucom.2019.11.023}}.
\newline\urlprefix\url{https://doi.org/10.1016/j.neucom.2019.11.023}

\bibitem{deng2009imagenet}
J.~Deng, W.~Dong, R.~Socher, L.-J. Li, K.~Li, L.~Fei-Fei, Imagenet: A
  large-scale hierarchical image database, in: 2009 IEEE conference on computer
  vision and pattern recognition, Ieee, 2009, pp. 248--255.

\bibitem{krizhevsky2012imagenet}
A.~Krizhevsky, I.~Sutskever, G.~E. Hinton, Imagenet classification with deep
  convolutional neural networks, in: Advances in neural information processing
  systems, 2012, pp. 1097--1105.

\bibitem{DBLP:journals/corr/SimonyanZ14a}
K.~Simonyan, A.~Zisserman, \href{http://arxiv.org/abs/1409.1556}{Very deep
  convolutional networks for large-scale image recognition}, in: Y.~Bengio,
  Y.~LeCun (Eds.), 3rd International Conference on Learning Representations,
  {ICLR} 2015, San Diego, CA, USA, May 7-9, 2015, Conference Track Proceedings,
  2015.
\newline\urlprefix\url{http://arxiv.org/abs/1409.1556}

\bibitem{he2016deep}
K.~He, X.~Zhang, S.~Ren, J.~Sun, Deep residual learning for image recognition,
  in: Proceedings of the IEEE conference on computer vision and pattern
  recognition, 2016, pp. 770--778.

\bibitem{he2016identity}
K.~He, X.~Zhang, S.~Ren, J.~Sun, Identity mappings in deep residual networks,
  in: European conference on computer vision, Springer, 2016, pp. 630--645.

\bibitem{everingham2010pascal}
M.~Everingham, L.~Van~Gool, C.~K. Williams, J.~Winn, A.~Zisserman, The pascal
  visual object classes (voc) challenge, International journal of computer
  vision 88~(2) (2010) 303--338.

\bibitem{DBLP:journals/cacm/KrizhevskySH17}
A.~Krizhevsky, I.~Sutskever, G.~E. Hinton, Imagenet classification with deep
  convolutional neural networks, Commun. {ACM} 60~(6) (2017) 84--90.

\bibitem{ren2015faster}
S.~Ren, K.~He, R.~Girshick, J.~Sun, {Faster R-CNN: Towards real-time object
  detection with region proposal networks}, in: Advances in neural information
  processing systems, 2015, pp. 91--99.

\bibitem{DBLP:conf/cvpr/GirshickDDM14}
R.~B. Girshick, J.~Donahue, T.~Darrell, J.~Malik, Rich feature hierarchies for
  accurate object detection and semantic segmentation, in: 2014 {IEEE}
  Conference on Computer Vision and Pattern Recognition, {CVPR} 2014, Columbus,
  OH, USA, June 23-28, 2014, 2014, pp. 580--587.

\bibitem{DBLP:conf/iccv/Girshick15}
R.~B. Girshick, Fast {R-CNN}, in: 2015 {IEEE} International Conference on
  Computer Vision, {ICCV} 2015, Santiago, Chile, December 7-13, 2015, 2015, pp.
  1440--1448.

\bibitem{DBLP:journals/pami/RenHG017}
S.~Ren, K.~He, R.~B. Girshick, J.~Sun, Faster {R-CNN:} towards real-time object
  detection with region proposal networks, {IEEE} Trans. Pattern Anal. Mach.
  Intell. 39~(6) (2017) 1137--1149.

\bibitem{DBLP:conf/cvpr/RedmonDGF16}
J.~Redmon, S.~K. Divvala, R.~B. Girshick, A.~Farhadi, You only look once:
  Unified, real-time object detection, in: 2016 {IEEE} Conference on Computer
  Vision and Pattern Recognition, {CVPR} 2016, Las Vegas, NV, USA, June 27-30,
  2016, 2016, pp. 779--788.

\bibitem{DBLP:conf/eccv/LiuAESRFB16}
W.~Liu, D.~Anguelov, D.~Erhan, C.~Szegedy, S.~E. Reed, C.~Fu, A.~C. Berg,
  {SSD:} single shot multibox detector, in: Computer Vision - {ECCV} 2016 -
  14th European Conference, Amsterdam, The Netherlands, October 11-14, 2016,
  Proceedings, Part {I}, 2016, pp. 21--37.

\bibitem{DBLP:journals/corr/abs-1804-02767}
J.~Redmon, A.~Farhadi, \href{http://arxiv.org/abs/1804.02767}{Yolov3: An
  incremental improvement}, CoRR abs/1804.02767.
\newblock \href {http://arxiv.org/abs/1804.02767} {\path{arXiv:1804.02767}}.
\newline\urlprefix\url{http://arxiv.org/abs/1804.02767}

\bibitem{MOTChallenge2015}
L.~Leal-Taix\'{e}, A.~Milan, I.~Reid, S.~Roth, K.~Schindler,
  \href{http://arxiv.org/abs/1504.01942}{{MOTC}hallenge 2015: {T}owards a
  benchmark for multi-target tracking}, arXiv:1504.01942 [cs]ArXiv: 1504.01942.
\newline\urlprefix\url{http://arxiv.org/abs/1504.01942}

\bibitem{MOT19_CVPR}
P.~Dendorfer, H.~Rezatofighi, A.~Milan, J.~Shi, D.~Cremers, I.~Reid, S.~Roth,
  K.~Schindler, L.~Leal-Taix\'{e},
  \href{http://arxiv.org/abs/1906.04567}{{CVPR19} tracking and detection
  challenge: {H}ow crowded can it get?}, arXiv:1906.04567 [cs]ArXiv:
  1906.04567.
\newline\urlprefix\url{http://arxiv.org/abs/1906.04567}

\bibitem{MOTChallenge20}
P.~Dendorfer, H.~Rezatofighi, A.~Milan, J.~Shi, D.~Cremers, I.~Reid, S.~Roth,
  K.~Schindler, L.~Leal-Taix\'{e},
  \href{http://arxiv.org/abs/1906.04567}{Mot20: A benchmark for multi object
  tracking in crowded scenes}, arXiv:2003.09003[cs]ArXiv: 2003.09003.
\newline\urlprefix\url{http://arxiv.org/abs/1906.04567}

\bibitem{DBLP:conf/icip/BewleyGORU16}
A.~Bewley, Z.~Ge, L.~Ott, F.~T. Ramos, B.~Upcroft,
  \href{https://doi.org/10.1109/ICIP.2016.7533003}{Simple online and realtime
  tracking}, in: 2016 {IEEE} International Conference on Image Processing,
  {ICIP} 2016, Phoenix, AZ, USA, September 25-28, 2016, {IEEE}, 2016, pp.
  3464--3468.
\newblock \href {http://dx.doi.org/10.1109/ICIP.2016.7533003}
  {\path{doi:10.1109/ICIP.2016.7533003}}.
\newline\urlprefix\url{https://doi.org/10.1109/ICIP.2016.7533003}

\bibitem{DBLP:conf/icip/WojkeBP17}
N.~Wojke, A.~Bewley, D.~Paulus,
  \href{https://doi.org/10.1109/ICIP.2017.8296962}{Simple online and realtime
  tracking with a deep association metric}, in: 2017 {IEEE} International
  Conference on Image Processing, {ICIP} 2017, Beijing, China, September 17-20,
  2017, {IEEE}, 2017, pp. 3645--3649.
\newblock \href {http://dx.doi.org/10.1109/ICIP.2017.8296962}
  {\path{doi:10.1109/ICIP.2017.8296962}}.
\newline\urlprefix\url{https://doi.org/10.1109/ICIP.2017.8296962}

\bibitem{DBLP:conf/wacv/WojkeB18}
N.~Wojke, A.~Bewley, \href{https://doi.org/10.1109/WACV.2018.00087}{Deep cosine
  metric learning for person re-identification}, in: 2018 {IEEE} Winter
  Conference on Applications of Computer Vision, {WACV} 2018, Lake Tahoe, NV,
  USA, March 12-15, 2018, {IEEE} Computer Society, 2018, pp. 748--756.
\newblock \href {http://dx.doi.org/10.1109/WACV.2018.00087}
  {\path{doi:10.1109/WACV.2018.00087}}.
\newline\urlprefix\url{https://doi.org/10.1109/WACV.2018.00087}

\bibitem{DBLP:conf/clear/2006}
R.~Stiefelhagen, J.~S. Garofolo (Eds.),
  \href{https://doi.org/10.1007/978-3-540-69568-4}{Multimodal Technologies for
  Perception of Humans, First International Evaluation Workshop on
  Classification of Events, Activities and Relationships, {CLEAR} 2006,
  Southampton, UK, April 6-7, 2006, Revised Selected Papers}, Vol. 4122 of
  Lecture Notes in Computer Science, Springer, 2007.
\newblock \href {http://dx.doi.org/10.1007/978-3-540-69568-4}
  {\path{doi:10.1007/978-3-540-69568-4}}.
\newline\urlprefix\url{https://doi.org/10.1007/978-3-540-69568-4}

\bibitem{DBLP:conf/clear/2007}
R.~Stiefelhagen, R.~Bowers, J.~G. Fiscus (Eds.),
  \href{https://doi.org/10.1007/978-3-540-68585-2}{Multimodal Technologies for
  Perception of Humans, International Evaluation Workshops {CLEAR} 2007 and
  {RT} 2007, Baltimore, MD, USA, May 8-11, 2007, Revised Selected Papers}, Vol.
  4625 of Lecture Notes in Computer Science, Springer, 2008.
\newblock \href {http://dx.doi.org/10.1007/978-3-540-68585-2}
  {\path{doi:10.1007/978-3-540-68585-2}}.
\newline\urlprefix\url{https://doi.org/10.1007/978-3-540-68585-2}

\end{thebibliography}

\end{document}